\documentclass[sigconf]{acmart}
\AtBeginDocument{%
  }

\setcopyright{acmlicensed}
\copyrightyear{2018}
\acmYear{2018}
\acmDOI{XXXXXXX.XXXXXXX}
\acmISBN{978-1-4503-XXXX-X/2018/06}

\usepackage[dvipsnames]{xcolor}
\usepackage{tcolorbox}
\usepackage{caption}
\usepackage{subcaption}
\usepackage{mdframed}
\tcbuselibrary{breakable}
\usepackage[T1]{fontenc}
\usepackage{booktabs} 
\usepackage{multirow} 
\usepackage{siunitx}  

\usepackage{listings}
\usepackage{graphicx}
\usepackage{xcolor}

\begin{document}

\title{Leveraging Foundation Models for Causal Generative Modeling}

\author{Aneesh Komanduri}
\affiliation{%
  \institution{University of Arkansas}
  \city{Fayetteville}
  \state{Arkansas}
  \country{USA}
}
\email{akomandu@uark.edu}

\author{Xintao Wu}
\affiliation{%
    \institution{University of Arkansas}
  \city{Fayetteville}
  \state{Arkansas}
  \country{USA}
  }
\email{xintaowu@uark.edu}

\renewcommand{\shortauthors}{Komanduri et al.}

\begin{abstract}
    Causal generative modeling is essential for developing reliable and transparent AI systems capable of counterfactual reasoning. While existing approaches focus on integrating causal constraints during the training of generative models, they often lack a unified framework to leverage the zero-shot reasoning capabilities of pretrained foundation models. We introduce FM-CGM, a modular framework for end-to-end visual causal reasoning using pretrained foundation models. FM-CGM formalizes the causal pipeline through three core components: a concept extractor, a concept manipulator, and a counterfactual generator. By leveraging a large reasoning model for causal inference and a text-to-image diffusion model for generation, our approach enables zero-shot causal discovery, intervention, and counterfactual generation. We then develop Causal Semantic Guidance (CSG), a cross-attention-based mechanism that ensures semantic interventions propagate to descendant concepts while preserving invariant regions. We empirically show that our approach can identify plausible causal structures and is suitable for faithful counterfactual image generation.
\end{abstract}





\received{20 February 2007}
\received[revised]{12 March 2009}
\received[accepted]{5 June 2009}

\maketitle

\section{Introduction}
Causal generative modeling \cite{komanduri2024from} has been studied both in the context of discovering causal concepts and their relationships from high-dimensional data \cite{scholkopf_toward_2021} and reasoning about counterfactual scenarios \cite{ribeiro2025counterfactual}. This direction has contributed to improving the reliability and transparency of AI systems. Existing work in causal generative modeling primarily focuses on integrating causality directly into the training process of generative models such as GANs \cite{kocaoglu2017causalgan}, VAEs \cite{deepscm, komanduri2024learning}, and Diffusion models \cite{komanduri2024causaldiffae} by modeling relationships among generative causal factors. Several works have also studied counterfactual image generation using state-of-the-art diffusion models \cite{sanchez2022diffusion, spyrou2025causally}. Recently, due to the inference capabilities of large-scale generative models such as large language model (LLMs) and large reasoning models, there has been ongoing research utilizing the power of foundation models as domain experts to improve causal discovery and inference \cite{vashishtha2023causal}, and causal agent frameworks to automate tasks in scientific discovery \cite{verma2025causal}. However, there is no unified framework that leverages the capabilities of pretrained foundation models to discover generative factors, reason about interventions, and generate counterfactual scenarios, the complete pipeline of causal generative modeling. 

In this work, we propose a formalism for visual causal reasoning from the perspective of pretrained foundation models. We develop \textit{Foundation Model Powered Causal Generative Model (FM-CGM)}, a high-level abstraction for causal generative modeling using pretrained foundation model components. We formalize the key components of a visual causal generative model in this setting as (1) \textit{concept extractor}, which infers a set of concepts and their causal relationships from a given image, (2) \textit{concept manipulator}, a mechanism to perform interventions on concept variables and propagate causal effects, and (3) a \textit{counterfactual generator}, a generation mechanism to map the concepts under intervention to a counterfactual image. 

We represent each component using a powerful large-scale foundation model. Specifically, we implement a practical method consisting of a reasoning VLM (Qwen3-VL) as the concept extractor and manipulator and text-to-image generative model (Stable Diffusion XL) as the counterfactual generator. To translate semantic interventions inferred by the concept manipulator into pixel-space counterfactual edits, we propose \textit{causal semantic guidance (CSG)} a procedure that operates through the text-to-image diffusion model's text-conditioning mechanism. Concretely, we leverage cross-attention attribution maps to localize the regions associated with each concept and its descendants and construct an inference-time guidance mechanism that (a) amplifies edits for intervened concepts, (b) encourages consistent changes in descendant concepts, and (c) suppresses unintended changes to non-descendant (invariant) concepts. This yields counterfactual images that are both visually plausible and aligned with the intended causal semantics.

\textbf{Contributions.} We summarize our contributions as follows (1) We propose a general and modular framework for end-to-end causal generative modeling with pretrained foundation model components including a \textit{concept extractor}, \textit{concept manipulator}, and \textit{counterfactual generator}, each represented as a foundation model. (2) We develop \emph{causal semantic guidance (CSG)}, a cross-attention-based guidance method that compositionally enforces the concept under intervention and its descendants, inferred by a large reasoning model, while preserving non-descendant concepts, enabling faithful counterfactual generation. (3) We demonstrate empirically that our method can perform minimal and faithful counterfactual edits.

\section{Related Work}

\paragraph{\textbf{Diffusion Models for Controllable Generation}.} Diffusion models \cite{ho_ddpm, improved_diffusion, song2021denoising} have quickly become one of the most effective tools for controllable image generation with models such as DALL-E \cite{pmlr-v139-ramesh21a}, StableDiffusion \cite{Rombach_2022_CVPR}, T2I-Adapter \cite{mou2024t2i}, ControlNet \cite{zhang2023adding}, etc. Several different techniques have been proposed to achieve controllable generation, such as fine-tuning, adaptation, and post-hoc optimization. Recent work in diffusion probabilistic models has focused on interpretable properties to enable generative control. \cite{wu_diffusiondisentanglement} show that pretrained text-to-image diffusion models, such as Stable Diffusion, already have somewhat of a content-style disentanglement capability, and propose a lightweight optimization to learn a soft mixing of content and style-infused prompts for better disentanglement. \cite{yang2023disdiff} study disentanglement properties of diffusion probabilistic models by learning separate gradient fields. \cite{brack2023Sega} propose LEDITS++, an image editing technique using cross-attention attribution maps capable of composing multiple concept edits. 

\paragraph{\textbf{Causality-based Diffusion Models}.} \cite{sanchez2022diffusion} provide a causal interpretation of classifier guidance in diffusion models and propose a method to guide the generation of image counterfactuals using interventions on labeled data. \cite{komanduri2024causaldiffae} explore counterfactual generation in diffusion models through guidance from disentangled causal representations. \cite{ribeiro2025counterfactual} study counterfactual identifiability in high-dimensions leveraging flow-matching. \cite{spyrou2025causally} develop a framework for counterfactual video editing in pretrained text-to-video generative models by utilizing VLMs to refine counterfactual prompts given a fixed causal graph. \cite{wang2023concept} formalize the notion of causal separability in text-to-image generative models through a counterfactual lens and show that concepts that are independently manipulable lie in orthogonal subspaces. Based on this intuition, they propose a simple projection mechanism to manipulate style components of generated images while preserving content. 

\paragraph{\textbf{Large Vision-Language Models}.} Extending LLMs to incorporate visual inputs, large vision-language models (LVLMs) have shown impressive performance in tasks such as recognition and visual question answering (VQA). Previous work has shown that LVLMs can sometimes struggle to perform formal causal reasoning \cite{komanduri-etal-2025-causalvlbench}. Recently, there has been extensive development in improving the reasoning abilities of LLMs and LVLMs through reinforcement learning techniques with verifiable rewards. Recently, there has been a large improvement in visual reasoning capabilities in vision language models, including QwenVL \cite{bai2025qwen3}, OpenAI o1, and Gemini-Thinking. These models have shown remarkable performance on math, coding, and visual reasoning tasks.

\section{Preliminaries}

\paragraph{\textbf{Structural Causal Model}. \cite{Pearl09}} A \textbf{structural causal model (SCM)} is defined as $\mathcal{C} = \langle C, U, F \rangle$, where $C$ is the set of semantic concepts, $U$ is the set of exogenous noise factors, and $F$ is a set of independent causal mechanisms of the form
\begin{equation}
    C_i = f_i (U_i, C_{\textbf{pa}_i})
\end{equation}
where $\forall i$, $f_i: U_i \times \prod_{j\in \textbf{pa}_i} C_j \to C_i$ are \textbf{causal mechanisms} that determine each causal variable as a function of the parents and noise, $C_{\textbf{pa}_i}$ are the parents of causal variable $C_i$.

\paragraph{\textbf{Diffusion Probabilistic Models}.} Diffusion models are a class of likelihood-based generative models that consist of two main steps: forward and reverse diffusion. Given an input image $X_0$ sampled from some data distribution, the \textit{forward diffusion} process defines a Markov chain of diffusion steps to slowly destroy the structure of data through a series of Gaussian noise perturbations defined as follows:
\begin{equation}
    X_t = \sqrt{\bar{\alpha_t}}X_0 + (\sqrt{1-\bar{\alpha}_t})\epsilon
\end{equation}
where $\epsilon \sim \mathcal{N}(\mathbf{0}, \mathbf{I})$, $t\sim \text{Unif}(0, T)$ and $\bar{\alpha_t} = \prod_{i=1}^t \alpha_i$. The \textit{reverse diffusion} restores the structure of data by parameterizing a UNet model $\epsilon_{\theta}$ to predict the noise perturbation at each time step $t$ given input image $X_t$ and minimizing the mean squared error between the predicted and actual noise at each time step $t$. During inference, random noise $X_T$ is sampled from a standard Gaussian distribution and $\epsilon_{\theta}$ is applied iteratively to denoise $X_T$ to a clean image $X_0$. The overall objective of the diffusion model is as follows
\begin{equation}
    \mathcal{L}_{\text{simple}} = \mathbb{E}_{X_t, t, \epsilon}(\|\epsilon - \epsilon_{\theta}(X_t, t)\|^2_2)
\end{equation}
Latent diffusion models (LDMs) apply the same procedure in the latent space of a well-trained variational autoencoder that compresses the image data $X_0$ to a perceptually equivalent low-dimensional representation $z_0$. Operating in the latent space has significant advantages such as faster training and sampling. The noise parameterization UNet $\epsilon_{\theta}$ is modified to perform denoising of the latent variables. Stable Diffusion (SD), a specific implementation of the latent diffusion model, is a \textit{text-to-image} generative model that conditions the reverse diffusion process on a text prompt $Y$ from a CLIP \cite{radford2021learning} text encoder. The prompt embedding influences the feature maps of the UNet through cross-attention mechanisms. The objective of Stable Diffusion is as follows
\begin{equation}
    \mathcal{L}_{\text{StableDiffusion}} = \mathbb{E}_{z_t, t, \epsilon}(\|\epsilon - \epsilon_{\theta}(z_t, t, Y)\|^2_2)
\end{equation}

\section{Foundation Model Powered Causal Generative Model}

\begin{figure*}
    \centering
    \includegraphics[width=\linewidth]{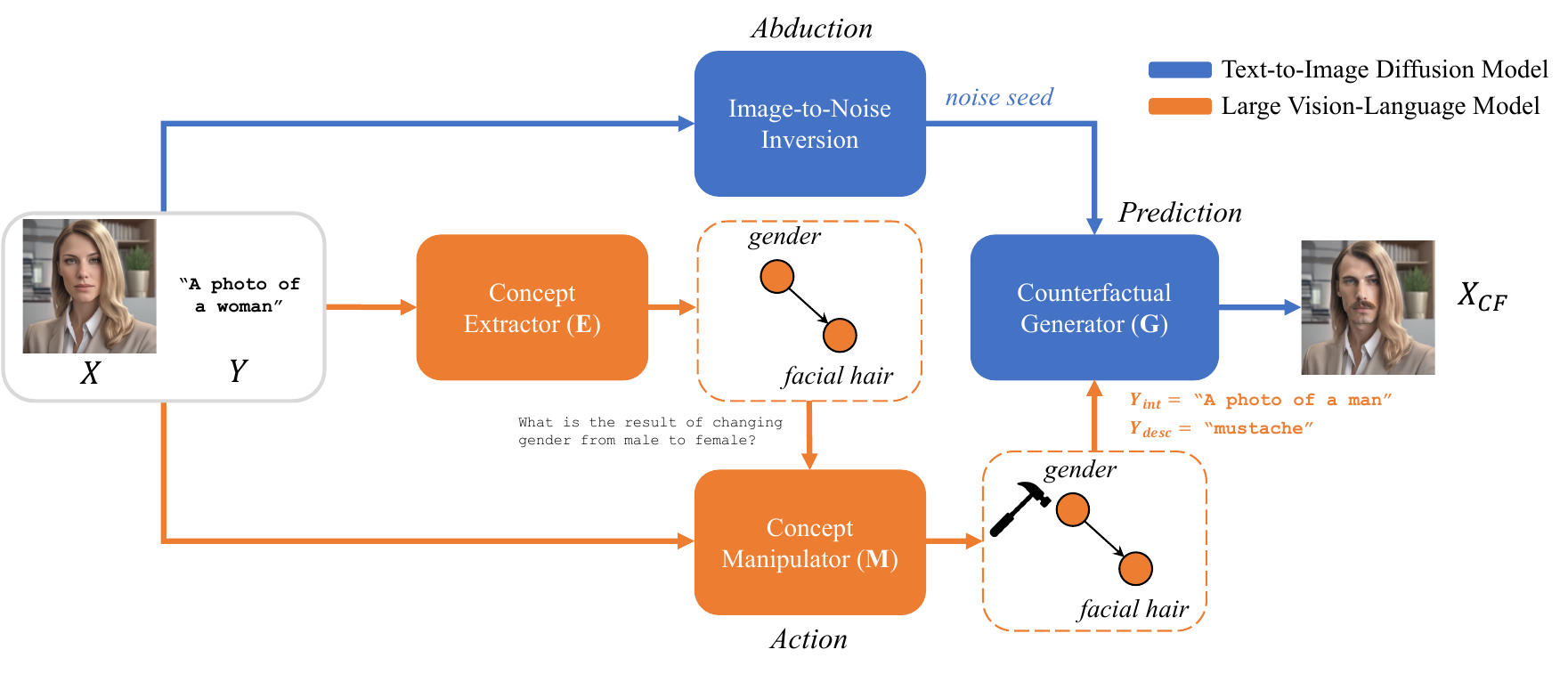}
    \caption{An overview of \textit{Foundation Model Powered Causal Generative Model (FM-CGM)} consisting of a concept extractor, concept manipulator, and counterfactual generator enabled by foundation models}
    \label{fig:arch}
\end{figure*}
In order to define a causal generative model, we require three components: (1) a mechanism to infer the semantic concepts and their causal relationships from an image (\textit{concept extractor}), (2) an intervention mechanism to manipulate concepts and their descendants (\textit{concept manipulator}), and (3) a generation mechanism to map intervened concepts to a counterfactual image (\textit{counterfactual generator}). We describe each component in more detail and the desiderata for pretrained foundation models to simulate each component in the vision-language setting as follows:

\paragraph{\textbf{Concept Extractor}.} A \textit{concept extractor foundation model $\mathbf{E}$} maps an image $X$ and a text description $Y$ to a discrete set of textual concepts $C$ and a directed acyclic graph $\mathcal{G}$ describing the relationships among concepts as follows:
\begin{equation}
    (C, \mathcal{G}) = \mathbf{E}(X,Y)
    \label{eq:extractor}
\end{equation}
The concepts are assumed to be described by a semantic Structural Causal Model (SCM) $\mathcal{C}$, where (conceptually) the joint distribution for the causal concepts follows a Markov factorization: 
\begin{equation}
p(C_{1},...,C_{n})=\prod_{i=1}^{n}p(C_{i}|C_{\textbf{pa}_{i}})
\end{equation}

\paragraph{\textbf{Concept Manipulator}.} Given an inferred set of concepts and their causal relationships, a \textit{concept manipulator foundation model $\mathbf{M}$} performs an intervention on a concept and propagates effects to descendant concepts consistent with the causal graph $\mathcal{G}$
\begin{equation}
    C' = \mathbf{M}(X, Y, (C, \mathcal{G}), \textbf{do}(C_i=c_i'))
\label{eq:manipulator}
\end{equation}
where $C'$ is the set of concepts after intervention on concept $C_i$ and $c_i'$ is the specific value of the intervention. Specifically, this component performs a \textbf{do} intervention on a concept variable (e.g., \textbf{do}($C_i = \text{male}$) if $C_i$ describes the gender concept).

\paragraph{\textbf{Counterfactual Generator}.} A \textit{counterfactual generator foundation model $\mathbf{G}$} takes as input a set of concepts $C'$ after an intervention, and a representation of the factual image $X$ (for structural preservation), and generates a counterfactual image consistent with the intervention
\begin{equation}
    X_{CF} = \mathbf{G}(C', X)
    \label{eq:generator}
\end{equation} 
The generation mechanism maps concepts to measurable variables in the high-dimensional image (e.g., pixel groups representing a concept). That is, $X = g(X_{C_1}, \dots, X_{C_n}, \xi)$ where $g$ is one-to-one mapping describing a set of $X$-measurable concepts composing the generated image. Intuitively, the image $X$ is generated by abstract causal concept variables with causal mechanisms that are independent and modular with respect to the image. For example, in a text-to-image generative model, we can consider concepts to manifest in the image as high values in a region of the image's attention maps corresponding to the semantic concept.

\begin{figure*}[t]
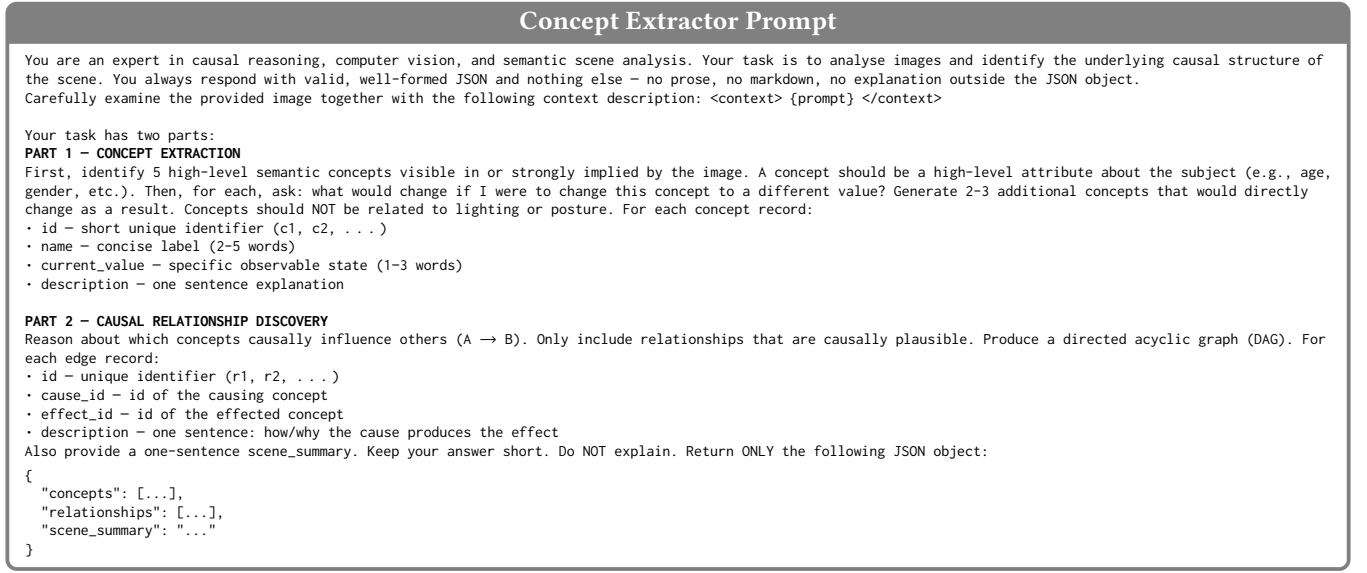

\centering
\begin{tcolorbox}[
    width=\textwidth, 
    colback=white, 
    colframe=gray, 
    fontupper=\ttfamily\scriptsize,
    boxsep=2pt, top=2pt, bottom=2pt, left=4pt, right=4pt, 
    title={\centering \textbf{\large Concept Extractor Prompt}}
]
\raggedright
You are an expert in causal reasoning, computer vision, and semantic scene analysis. Your task is to analyse images and identify the underlying causal structure of the scene.
You always respond with valid, well-formed JSON and nothing else — no prose, no markdown, no explanation outside the JSON object.

Carefully examine the provided image together with the following context description: <context>
\{prompt\}
</context> \\ [\baselineskip]

Your task has two parts: \\

\textbf{PART 1 — CONCEPT EXTRACTION} \\
First, identify 5 high-level semantic concepts visible in or strongly implied by the image. A concept should be a high-level attribute about the subject (e.g., age, gender, etc.). Then, for each, ask: what would change if I were to change this concept to a different value? Generate 2-3 additional concepts that would directly change as a result. Concepts should NOT be related to lighting or posture. For each concept record: \\
• id — short unique identifier (c1, c2, …) \\
• name — concise label (2-5 words) \\
• current\_value — specific observable state (1–3 words) \\
• description — one sentence explanation \\ [\baselineskip]

\textbf{PART 2 — CAUSAL RELATIONSHIP DISCOVERY} \\
Reason about which concepts causally influence others (A $\rightarrow$ B). Only include relationships that are causally plausible. Produce a directed acyclic graph (DAG). For each edge record: \\
• id — unique identifier (r1, r2, …) \\
• cause\_id — id of the causing concept \\
• effect\_id — id of the effected concept \\
• description — one sentence: how/why the cause produces the effect \\

Also provide a one-sentence scene\_summary. Keep your answer short. Do NOT explain. Return ONLY the following JSON object: \\
\begin{tcbverbatimwrite}{tmp.json}
{
  "concepts": [
    {"id": "c1", "name": "...", "current_value": "...", "description": "..."}
  ],
  "relationships": [
    {"id": "r1", "cause_id": "c1", "effect_id": "c2", "description": "..."}
  ],
  "scene_summary": "..."
}
\end{tcbverbatimwrite}
\begin{verbatim}
{
  "concepts": [...],
  "relationships": [...],
  "scene_summary": "..."
}
\end{verbatim}
\end{tcolorbox}
\caption{Concept extractor prompt}
\label{prompt:extractor}
\end{figure*}

\paragraph{\textbf{Foundation Model Powered Causal Generative Model}.} Formally, we define a class of visual causal generative models with pretrained foundation model components as follows:
\begin{definition}
An \textbf{Foundation Model Powered Causal Generative Model (FM-CGM)} is a quadruplet $\mathbf{F} = \langle \mathcal{C}, \mathbf{E}, \mathbf{M}, \mathbf{G}\rangle$ where:
\begin{itemize}
\item $\mathcal{C} = (C, \mathcal{G})$ is a semantic structural causal model describing a set of high-level concepts $C$ and their causal relationships $\mathcal{G}$ corresponding to the image \footnote{A semantic structural causal model is one that only includes high-level concepts and their relationships and not any causal mechanisms to comply with our setting}
\item $\mathbf{E}$ is a concept extractor represented by a large reasoning foundation model to infer a set of concepts $C$ and their causal relationships $\mathcal{G}$, as described in Eq. \ref{eq:extractor}
\item $\mathbf{M}$ is a concept manipulator represented by a large reasoning foundation model to reason about interventions on causal concepts (e.g., $\textbf{do}(C_i=\text{male})$), as described in Eq. \ref{eq:manipulator}
\item $\mathbf{G}$ is a counterfactual generator represented by a large text-to-image generative foundation model to map interventions on high-level concepts to a counterfactual image, as described in Eq. \ref{eq:generator}
\end{itemize}
\end{definition}

\section{Methodology}
In this section, we propose a practical foundation model-powered architecture to implement the FM-CGM, consisting of a Vision Language Model (VLM) to represent the concept extractor and concept manipulator and a text-to-image diffusion model (i.e., Stable Diffusion) to represent the counterfactual generator. The overall FM-CGM framework is shown in Figure \ref{fig:arch}.

\subsection{VLM-based Concept Extractor}
In this section, we provide the details of the \textit{Concept Extractor}. We represent the concept extractor as a pretrained reasoning vision-language model (VLM) that maps an image to a textual description of a concept. A concept variable is a categorical variable that takes exactly one value at a time and describes the visual scene. The resulting (text) concept serves as the semantic state for downstream counterfactual generation.

We first set up our motivation behind using an VLM to extract causally related concepts. \cite{zecevic2023causal} conjecture that large-scale foundation models learn \textit{causal facts}, but not physical mechanisms. We rely on this assumption to obtain a plausible causal structure over identified concepts (a \textit{meta-SCM}) from a reasoning model, which is sufficient for constraining downstream counterfactual edits. The diffusion decoder (via guidance) then provides the image-level mechanism that realizes these semantic changes.

We first detail a multi-stage approach to identifying useful causally related concepts by leveraging the reasoning abilities of large reasoning vision language models. We construct a robust prompt to identify the most important elements in a visual scene and their relationships as follows
\begin{enumerate}
    \item \textbf{Visual Description:} We prompt the VLM to describe the visual scene for context to obtain a base prompt $Y$.
    \item \textbf{Concept Identification:} We prompt the VLM to return a list of the most important and interpretable concepts (nodes) that describe the visual scene. We define this output as a concept set $C=\{C_1,\ldots,C_n\}$ for the given image where a concept variable $C_i$ takes a specific valuation specified by $c_i$.
    \item \textbf{Causal Graph:} Given the identified set of concepts $C$, we prompt the VLM to identify edges between pairs of concepts ($C_i$, $C_j$) if $C_i$ is a direct cause of $C_j$ to generate a causal graph $\mathcal{G}$.
\end{enumerate}
We obtain a set of fully specified concepts $C$ along with a causal graph $\mathcal{G}$ with which we can reason about causal queries. The complete prompt template for VLM concept and graph identification is specified in Figure~\ref{prompt:extractor}. For visual scenes, we find it reasonable to leverage strong reasoning-capable VLMs to output consistent concept sets and plausible causal graphs.

\begin{figure*}[t]
\centering
\begin{tcolorbox}[
    width=\textwidth, 
    colback=white, 
    colframe=gray, 
    fontupper=\ttfamily\scriptsize, 
    title={\centering \textbf{\large Concept Manipulator Prompt}}
]
\raggedright
You are an expert at understanding visual scenes and reasoning about how changing one thing in a scene affects other things. You always respond with valid, well-formed JSON and nothing else — no prose, no markdown, no explanation outside the JSON object. You are given an image, a description of the scene it shows, a list of concepts present in it, and a list of relationships that describe how those concepts influence one another. \\[\baselineskip]

<scene\_summary>
\{scene\_summary\}
</scene\_summary> 

<concepts>
\{concepts\_json\}
</concepts> 

<relationships>
\{relationships\_json\}
</relationships> \\ [\baselineskip]

Your task: \\

\textbf{1. PROPOSE 3 DIVERSE MANIPULATIONS} \\
• Each manipulation changes a DIFFERENT concept to a new value. \\
• Manipulations should produce visually distinct scenes. \\
• Pick one concept that affects many others, one that affects only a few, and one that affects none. \\
• Keep each manipulation realistic and physically plausible. \\[\baselineskip]

\textbf{2. WORK OUT WHAT ELSE CHANGES} \\
For each manipulation, think through every concept that would be different as a result of the changed concept. For each concept that changes, write its new value in 1–3 descriptive, unambiguous words that combine the concept variable and value (e.g. "wet road"). DO NOT INCLUDE CONCEPTS THAT STAY THE SAME. \\[\baselineskip]

\textbf{3. LIST FINAL STATE OF ALL CONCEPTS THAT CHANGED} \\
Record the final value of every concept that changed. Each value must combine the concept variable and its new value into 1–3 words (e.g., "black hair"). \\[\baselineskip]

\textbf{4. WRITE AN IMAGE GENERATION PROMPT} \\
Write a generation\_prompt that adds to or modifies the following reference prompt: 
<reference\_prompt> \{base\_prompt\} </reference\_prompt> \\[\baselineskip]

Return ONLY the following JSON: 
\begin{verbatim}
{
  "interventions": [{"id": "intervention_1", "target_concept_id": "...", "target_concept_name": "...", "intervention_description": "...", "original_value": "...", 
  "new_value": "...",
      "propagated_changes": [{"concept_id": "...", "concept_name": "...", "original_value": "...", "new_value": "...", "reason": "..."}],
      "final_concept_states": { "c1": "value", "c2": "value" },
      "generation_prompt": "..."
    }
  ]
}
\end{verbatim}
\end{tcolorbox}
\caption{Concept manipulator prompt}
\label{prompt:manipulator}
\end{figure*}

\subsection{VLM-based Concept Manipulator} 

Define $I = \{1, \dots, n\}$ as the indices for concepts. For an intervention target $i\in I$ with intervention $\textbf{do}(C_i=c_i')$ on a causal concept, we prompt the VLM to obtain a counterfactual concept set $C'$ such that (1) the intervention $\textbf{do}(C_i=c_i')$ is performed, (2) $\textbf{De}(C_i)$ (descendants) are updated to values consistent with the intervention and the causal graph, and (3) $\textbf{ND}(C_i)$ (non-descendants) are preserved. This semantic counterfactual state $C'$ is then translated into diffusion guidance as defined below. The prompt for the concept manipulator is specified in Figure~\ref{prompt:manipulator}.

\subsection{Diffusion-based Counterfactual Generator}

In this section, we provide the details of the \textit{Counterfactual Generator}. Since we cannot mechanistically change the Stable Diffusion model to enforce structural causal mechanisms in the latent space, we primarily rely on guidance satisfying \textit{counterfactual minimality}. In this section, we take inspiration from the LEDITS++ framework \cite{brack2023Sega} and develop a VLM-informed counterfactual generation method \textbf{Causal Semantic Guidance (CSG)}, a classifier-free guidance approach coupled with a VLM as the \textit{Concept Manipulator} to achieve faithful counterfactual image generation. Following the Pearlian paradigm \cite{Pearl09}, we implement the counterfactual edit through a structured Abduction-Action-Prediction style procedure.

\paragraph{\underline{\textbf{Abduction}}} We first capture the noise $z_T$ specific to the factual image $X_0$ by performing an inversion of the sampling process. We utilize the fast second-order DPM-Solver++ inversion \cite{lu2025dpm} to obtain the latent representation $z_T$ via forward diffusion.

\paragraph{\underline{\textbf{Action}}} Given the counterfactual states $C'$ obtained from the VLM Concept Manipulator, we can now utilize classifier-free guidance and disentangled image editing techniques to realize counterfactual image edits. Now, for a given intervened concept $C_{i}$, we construct an interventional prompt $Y_{C_{i}}$ following the style of the base prompt $Y$ and a set of prompts $Y_{\textbf{De}(C_i)}$ for descendants. For the collection of descendant concepts, we define the guidance term as the following sum
\begin{equation}
    \gamma(X_t, Y_{\textbf{De}(C_i)}) = \sum_{c\in \textbf{De}(C_i)} \phi(\psi; s_{c}, \lambda) \psi(X_t, Y_c)
\end{equation}
where $\phi$ applies an elementwise edit guidance scale $s_c$, and $\psi$ is defined as
\begin{equation}
    \psi(X_t, Y_c) = \epsilon_{\theta}(X_t, Y_c) - \epsilon_{\theta}(X_t)
\end{equation}
The $\phi$ term scales elements of the image and its score estimate that are relevant to the prompt $Y$. A larger scaling factors $s_c$ naturally increases the effect of the intervention, and $\lambda$ selects the relevant pixels. 
The term $\phi$ is an intersection of a binary masks $M^1_c$ generated from UNet cross-attention layers for a concept $c$ and $M^2$ generated from the unconditioned noise estimate (thresholded by the value $\lambda$), scaled by a factor $s_c$. 
\begin{equation}
    \phi(\psi; s_{c}, \lambda) = s_{c}M_c^1 M^2
\end{equation}
The intersection of the masks focuses both on relevant image regions and fine-grained semantic information from the descendants. Intuitively, the $M^2$ preserves other features, including the non-descendant concepts $\textbf{ND}(C_i)$.

\begin{figure*}[t!]
    \centering
    \begin{subfigure}[b]{0.45\linewidth}
        \centering
        \includegraphics[width=\linewidth]{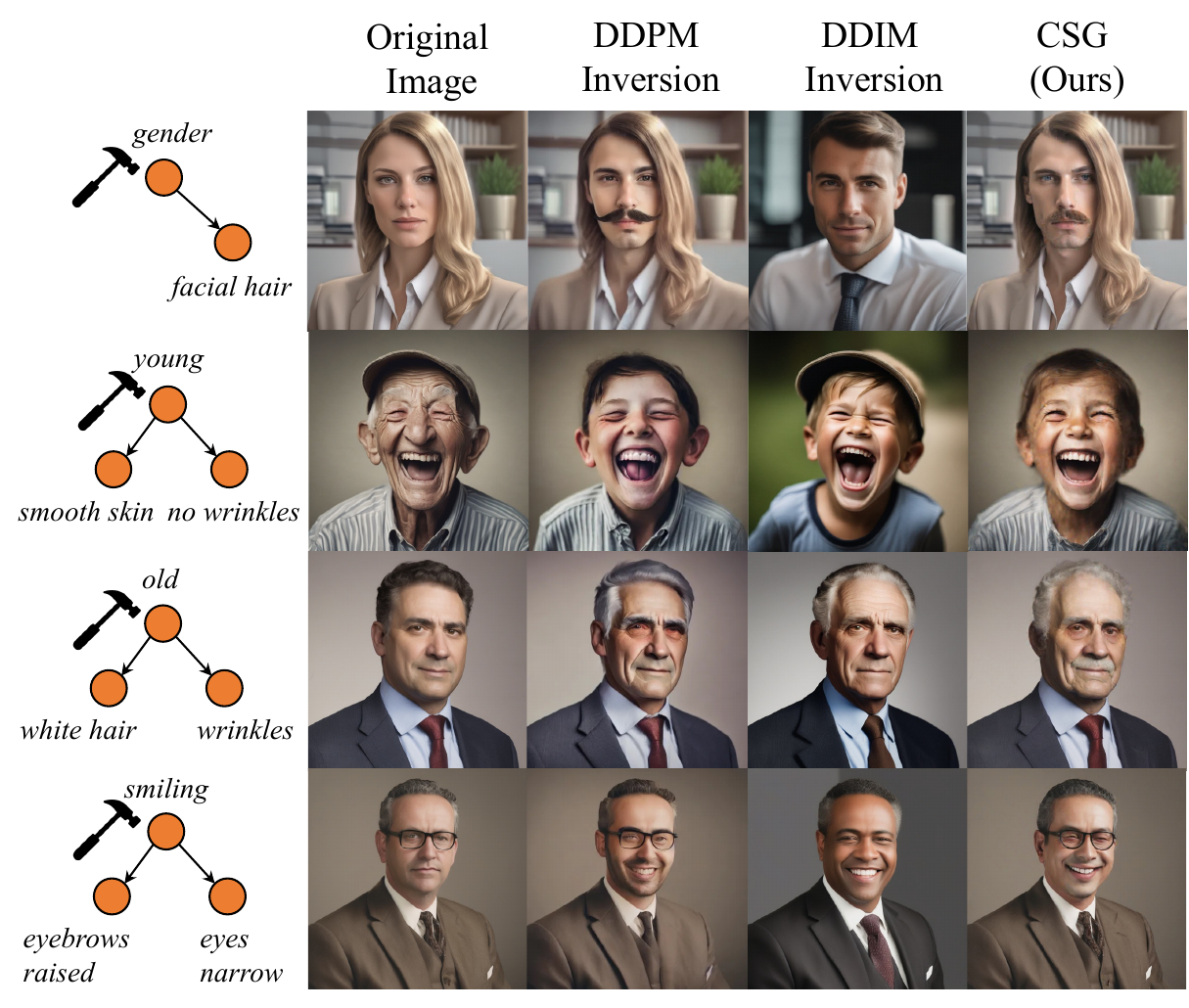}
        \caption{}
        \label{fig:exp1}
    \end{subfigure}~ 
    \begin{subfigure}[b]{0.45\linewidth}
        \centering
        \includegraphics[width=\linewidth]{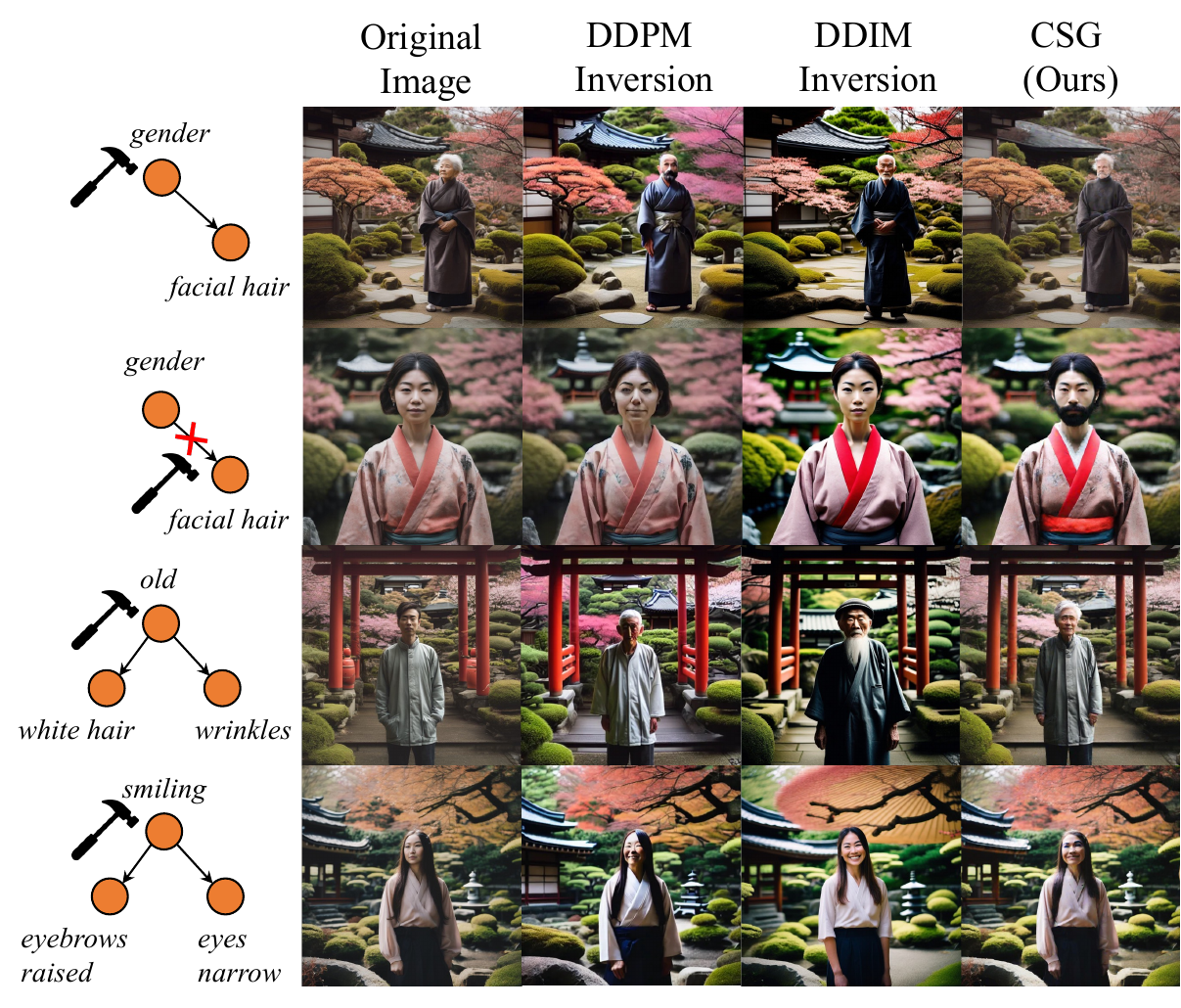}
        \caption{}
        \label{fig:exp2}
    \end{subfigure}
    \caption{Image Counterfactual Generation on human facial characteristics for (a) close-up profile and (b) garden scene}
    \label{fig:exp1_2}
\end{figure*}

\begin{figure}
    \centering
    \includegraphics[width=\linewidth]{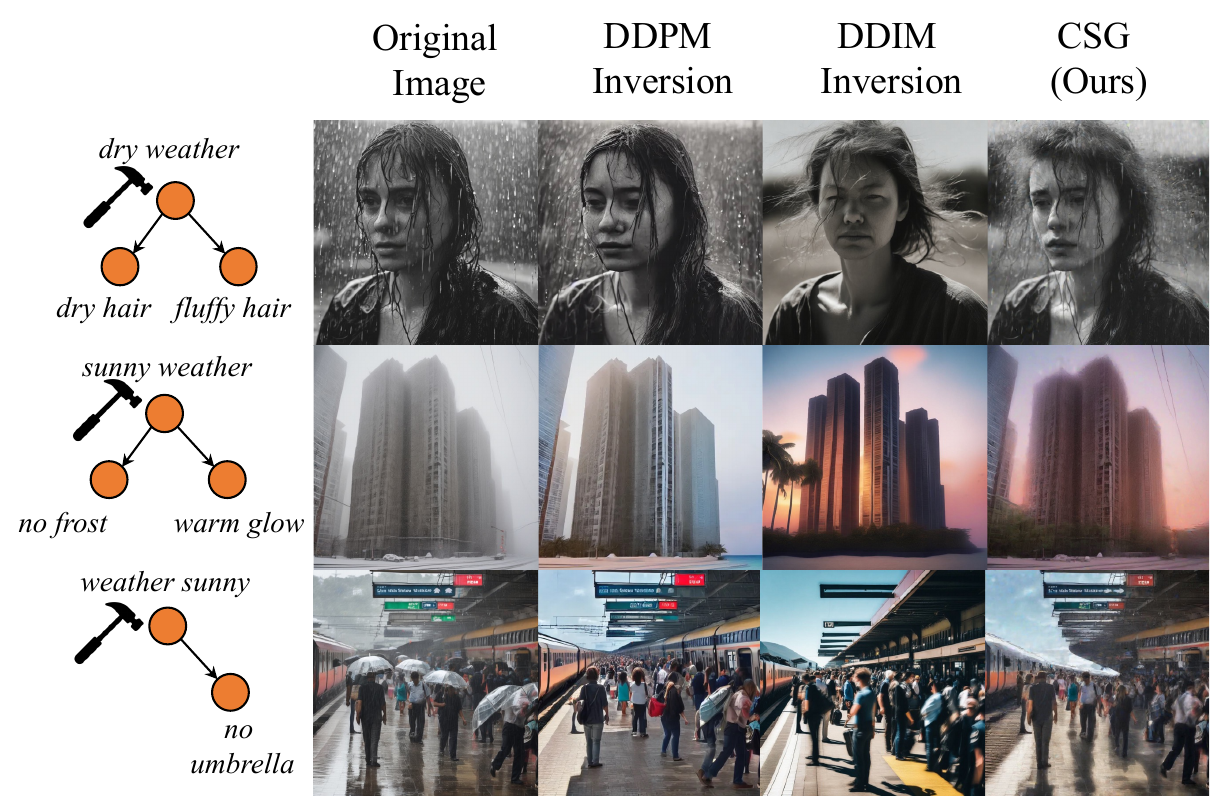}
    \caption{Counterfactual Generation on weather scenes}
    \label{fig:weather_exp}
\end{figure}

\paragraph{\underline{\textbf{Prediction}}} We define the score update as follows
\begin{equation}
\begin{split}
    \epsilon_{\theta}(X_t, Y_{C'}) = \epsilon_{\theta}(X_t, Y_{C_i}) + \gamma_t(X_t, Y_{\textbf{De}(C_i)}) 
\end{split}
\end{equation}
where the guidance term $\gamma$ pushes the interventional score estimate $\epsilon(X_t, Y_{C_i})$ toward the descendant concepts estimate to ultimately realize a minimal counterfactual edit satisfying the causal graph.

\section{Experiments}
In this section, we empirically show as a demonstration of concept that we can produce accurate image counterfactuals using our proposed framework. Specifically, we show the result of VLM interventional reasoning coupled with the compositional editing technique compared to baseline editing methods.

\paragraph{\textbf{Datasets}.} We qualitatively evaluate our method on a small set of images generated directly from Stable Diffusion XL (SDXL), as shown in Figures \ref{fig:exp1_2} and \ref{fig:weather_exp}. We also perform a larger scale experiment on two benchmark datasets: MS-COCO \cite{mscoco} and CelebA-HQ \cite{karras2018progressive}. MS-COCO consists of images with natural language captions describing diverse everyday scenes such as objects, animals, outdoor environments, and complex multi-entity compositions. CelebA-HQ is a high-quality human face dataset of 30,000 celebrity portrait images, each annotated with a descriptive text caption covering facial attributes such as age, gender, hair color, and expression. For each dataset, we randomly sample $75$ images per dataset using a fixed seed to ensure reproducibility, for a total of $N=150$ examples. Sampling is performed without replacement over the full available split. Each sampled image and its associated caption are passed through the complete pipeline: concept extraction, manipulation, generation across all editing methods.

\paragraph{\textbf{Implementation}.} Our pipeline implements automated causal concept discovery and counterfactual generation across four distinct stages. First, a Qwen3-VL-30B-A3B-Instruct \cite{bai2025qwen3} vision-language model (VLM) analyzes an input image and a textual description to extract a set of high-level semantic concepts, such as hair color or age, and their downstream effects. This stage constructs a directed acyclic graph (DAG) of causal relationships, with inference performed at a temperature of 0.1 and a budget of 2,048 tokens. Second, the VLM, conditioned on both the image and the discovered graph, proposes three diverse causal interventions at a temperature of 0.2 with a limit of 3,072 output tokens. Each intervention targets a unique concept, propagates its effects through the DAG, records all modified concept states as descriptive phrase-value pairs, and produces a text-to-image generation prompt.

Third, counterfactual images are generated with Stable Diffusion XL \cite{podell2024sdxl} via DDIMInversion, DDPMInversion as baselines, and Causal Semantic Guidance. For DDIMInversion and DDPMInversion, we incorporate classifier-free guidance with the fully specified prompt with all concepts included. For CSG, we use the LEDITS++ pipeline and a DPM-Solver++ scheduler to invert the original image into noise over 100 steps with a source guidance scale of 3.5 and an inversion skip ratio of 0.15. The system then denoises the image under the CSG edit using the propagated concept values as per-concept editing prompts. This involves an edit guidance scale of 8.0, an edit threshold of 0.7, 10 warmup steps, and intersect masking, while falling back to the direct manipulation value if no downstream concepts are affected.

\begin{table*}[t]
    \centering
    \caption{Evaluation Results}
    \begin{tabular}{l cc @{\hspace{1em}} cc @{\hspace{1em}} cc}
        \toprule
        \multirow{2}{*}{\textbf{Method}} & \multicolumn{2}{c}{\textbf{CelebA-HQ}} & \multicolumn{2}{c}{\textbf{MS-COCO}} & \multicolumn{2}{c}{\textbf{Average}} \\
        \cmidrule(r){2-3} \cmidrule(r){4-5} \cmidrule{6-7}
        & VLM-Eff. $\uparrow$ & LPIPS $\downarrow$ & VLM-Eff. $\uparrow$ & LPIPS $\downarrow$ & VLM-Eff. $\uparrow$ & LPIPS $\downarrow$ \\
        \midrule
        DDIM Inversion       & 0.792 & 0.4836 & 0.741 & 0.4759 & 0.767 & 0.4798 \\
        DDPM Inversion       & 0.756 & 0.2244 & 0.655 & 0.2035 & 0.705 & 0.2139 \\
        CSG (Ours)                 & \textbf{0.854} & \textbf{0.1980} & \textbf{0.762} & \textbf{0.1750} & \textbf{0.808} & \textbf{0.1865} \\
        \bottomrule
    \end{tabular}
    \label{tab:dataset_results}
\end{table*}

\begin{figure}[t]
    \centering
\begin{tcolorbox}[
    width=\columnwidth, 
    colback=white, 
    colframe=gray, 
    fontupper=\ttfamily\scriptsize, 
    title={\centering \textbf{\large VLM Effectiveness Evaluation Prompt}}
]
\raggedright
You look at images and answer simple questions about what is shown in them. \\[\baselineskip]
You always respond with valid, well-formed JSON and nothing else — no prose, no markdown, no explanation outside the JSON object.

Look at the given image. It was generated with the following prompt:
<prompt>
\{generation\_prompt\}
</prompt>

The one thing that was changed to produce this image was: \\
\hspace*{1em}\{target\_concept\_name\}: \{original\_value\} $\rightarrow$ \{new\_value\} \\[\baselineskip]

Below is a checklist of concepts and the value each one should have in the image. \\[\baselineskip]

<checklist>
\{checklist\_text\}
</checklist> \\[\baselineskip]

Your task: \\[\baselineskip]

\textbf{1. CHECKLIST EVALUATION} \\
For each item in the checklist, answer "yes" if that value is clearly present in the image, or "no" if it is not. \\[\baselineskip]

\textbf{2. VERDICT ASSIGNMENT} \\
• "success" — all checklist items are "yes" \\
• "partial" — most checklist items are "yes" but some are wrong \\
• "failure" — the main changed concept is not visible, OR most checklist items are "no" \\[\baselineskip]

\textbf{3. REASONING} \\
Write one or two sentences explaining your verdict. \\[\baselineskip]

Return ONLY the following JSON: \\
\begin{verbatim}
{
  "intervention_id": "...",
  "concept_checks": [
      {
      "concept_name": "...", 
      "expected_value": "...", 
      "present": "yes|no"
      }
  ],
  "verdict": "success|partial|failure",
  "reasoning": "..."
}
\end{verbatim}
\end{tcolorbox}
\caption{VLM effectiveness evaluation prompt}
\label{prompt:evaluator}
\end{figure}

\paragraph{\textbf{Qualitative Evaluation}.} We evaluate our method based on several image samples generated from Stable Diffusion XL. We find that our approach consistently outperforms baselines in generating minimal counterfactually consistent edits based on the causal graph specified by the VLM. For example, in Figure~\ref{fig:exp1}, intervening on the gender attribute should cause facial hair to change while preserving every other attribute in the original image. DDIM and DDPM Inversion introduce unnecessary variations, while CSG generates a minimal and consistent edit with downstream effects. We also find that our method works well when manipulating more complex concepts such as weather. In Figure~\ref{fig:weather_exp}, the first row shows an intervention on the weather from rainy to dry weather. The edit generated by our method shows a clear background and dry hair, which are inferred according to the VLM (e.g., sunny weather causes dry hair and fluffy hair) whereas baselines either add high variations or fail to incorporate causal changes.

\paragraph{\textbf{Quantitative Evaluation}.} We also provide a quantitative evaluation on a subset of the CelebA-HQ and MS-COCO datasets. Given a counterfactual image, we evaluate \textbf{effectiveness} of the intervention by using \textbf{VLM-Eff}, a VLM-based approach to identify whether the generated image contains each counterfactual concept value, similar to \cite{spyrou2025causally}. We include the effectiveness prompt in the Figure~\ref{prompt:evaluator} for reference. To evaluate \textbf{minimality}, or how much of a minimal edit the generated image is, we utilize perceptual similarity to the original image, quantified using \textbf{LPIPS} with a VGG backbone. We report the average VLM-Eff. and LPIPS score for all methods in Table \ref{tab:dataset_results}. Our results demonstrate that CSG is capable of more accurate counterfactual edits, while satisfying minimality compared to standard editing techniques.

\section{Conclusion}
In this work, we propose Foundation Model Powered Causal Generative Model, an abstraction for causal generative modeling consisting of pretrained foundation model components. We then develop a framework using large vision-language models for concept extraction and manipulation and Stable Diffusion XL for counterfactual generation. We propose Causal Semantic Guidance, an inference-time editing approach extending on LEDITS++ to ensure interventions and propagation to descendant concepts are sufficiently represented in the counterfactually generated image. Qualitative and quantitative empirical evaluation shows that our framework is promising for modularizing and automating the process of causal generative modeling by utilizing the power of foundation models.

\section*{Acknowledgements}
This work is supported in part by National Science Foundation under awards 1910284, 1946391, 2147375, the National Institute of General Medical Sciences of National Institutes of
Health under award P20GM139768, and the
Arkansas Integrative Metabolic Research Center
at University of Arkansas.

\bibliographystyle{ACM-Reference-Format}
\bibliography{main}

\end{document}